\documentclass[conference]{IEEEtran}
\ifCLASSINFOpdf
  \DeclareGraphicsExtensions{.pdf,.jpeg,.png,.eps}
\else
\fi

\usepackage{epsfig}
\usepackage{color}
\usepackage{subfigure}
\usepackage{epsfig}
\usepackage{color}
\usepackage{subfigure}
\usepackage{amsmath}
\usepackage{multicol}
\usepackage{amssymb}
\usepackage[font=small,skip=0pt]{caption}

\begin{document}
%
\title{A new recurrent neural network based predictive model for Faecal Calprotectin analysis: A retrospective study}


%
\author{\IEEEauthorblockN{Zeeshan Khawar Malik \IEEEauthorrefmark{1},
Zain U. Hussain\IEEEauthorrefmark{5}, Ziad Kobti \IEEEauthorrefmark{3}, Charlie W. Lees \IEEEauthorrefmark{4}, Newton Howard \IEEEauthorrefmark{5} and Amir Hussain\IEEEauthorrefmark{1}  \IEEEauthorblockA{\IEEEauthorrefmark{1}School of Natural Sciences, University of Stirling,\\}
\IEEEauthorrefmark{3} School of Computer Science, University of Windsor, Ontario, Canada\\
\IEEEauthorrefmark{5} Nuffield Department of Surgical Sciences, University of Oxford, UK\\
\IEEEauthorblockA{\IEEEauthorrefmark{4}Gastrointestinal Unit, Western General Hospital, Edinburgh, UK\\
\IEEEauthorrefmark{1} ahu, zkm@cs.stir.ac.uk , 
\IEEEauthorrefmark{4}charlie.lees@ed.ac.uk, \IEEEauthorrefmark{3}kobti@uwindsor.ca and 
\IEEEauthorrefmark{5} zain.hussain, newton.howard@nds.ox.ac.uk \\
}
}}


\maketitle

\begin{abstract}

Faecal Calprotectin (FC) is a surrogate marker for intestinal inflammation, termed Inflammatory Bowel Disease (IBD), but not for cancer. In this retrospective study of 804 patients, an enhanced benchmark predictive model for analyzing FC is developed, based on a novel state-of-the-art Echo State Network (ESN), an advanced dynamic recurrent neural network which implements a biologically plausible architecture, and a supervised learning mechanism. The proposed machine learning driven predictive model is benchmarked against a conventional logistic regression model, demonstrating statistically significant performance improvements.

\end{abstract}

\section{Introduction}

IBD involves chronic inflammation of the digestive tract and primarily includes Ulcerative colitis (UC) and Crohn's disease. Both are chronic relapsing diseases and are characterized as alternating between a state of remission and being active. Over the past few decades, the incidence of IBD around the world has risen dramatically \cite{zain1}. Traditionally, a colonoscopy is used to identify and confirm bowel inflammation, and is considered to be an accurate diagnostic tool. The procedure itself is expensive, and limited in parts of the world, and its invasive nature can result in discomfort and adverse effects for patients. The provision of a cheap and non-invasive diagnostic test would facilitate for early referral and detection of IBD.

Faecal calprotectin (FC), a calcium and zinc binding cytosolic protein found in neutrophils \cite{zain2}, is a biomarker specific for intestinal inflammation \cite{zain2}. The presence of FC in faeces quantitively relates to neutrophil migration to the gastrointestinal tract, which is why studies have shown increased FC in the stool of patients with IBD.

In healthy individuals, the upper limit of FC in faeces is 50 'g/g. Two large meta-analyses, of prospective studies in which patients had suspected IBD, found FC to be a very useful marker for diagnosing IBD, and differentiating it from Irritable Bowel Syndrome. The pooled specificity and sensitivity of FC was shown to be up to 93 \% and 96 \% respectively \cite{zain3},  \cite{zain4}. A large retrospective study carried out in Edinburgh \cite{zain2} corroborated existing data which showed that FC reliably distinguishes between patients with functional disease and IBD. It also showed that in recent years, FC has been used accurately for young adults (16-50 years) to exclude intestinal inflammation, and reduce the need for invasive procedures, as a negative FC would suggest a lack of an organic GI disease and GI Inflammation. Older patients still require colonoscopy to exclude colorectal cancer.

Commonly used statistical prediction techniques such as linear regression models are not complex enough to produce reliable results as they have poor statistical stability. Hence predicting the frequency of IBD using these traditional predictive models is not accurate. \cite{zain5} Artificial neural network (ANN) is a novel model, inspired by the functioning of the human brain. It can be used to build non-linear statistical models which can handle complex biological systems, and recently, has been successfully introduced in clinical settings for pattern recognition and survival prediction \cite{zain5}. Studies have shown the ANN model to outperforms both logistic regression and linear discriminant models, with increased accuracy \cite{zain6}. 

In this paper we propose a new recurrent neural network predictive model based on the Echo State Network (ESN) for retrospective FC analysis, which has been benchmarked against a conventional logistic regression model. The rest of the paper is organised as follows: Section \ref{section2} introduces the Calprotectin dataset and the methods used in this study, Sections \ref{section3} and \ref{section4}  describe the two methods employed, specifically, logistic regression and ESN, and also present comparative experimental results and discussions. Finally, concluding remarks and future work suggestions are outlined in Section \ref{section5}.

\section{Calprotectin Dataset}
\label{section2}

The retrospective dataset comprises records of 804 patients, which was provided for this study, by Western General Hospital’s Gastro Entomology (GI) Unit. Since the original data was in an unnormalized form, some pre-processing was carried out which led to a selection of 53 independent variables (out of the available 58). The objective of this initial case study was to determine the factors predictive of reaching a primary composite end point. A tenfold cross-validation technique was employed, and the prediction accuracy of the new ESN predictor was benchmarked against a conventional logistic regression model. ROC analysis was also carried out to verify the results. Table \ref{logit} illustrates the enhanced comparative accuracy of the proposed machine learning approach in demonstrating the predictive effectiveness of selected variables, for reaching the primary composite endpoint.

In machine learning and neural network communities, several neural network models and kernel-based methods are applied to predictive modeling tasks, such as MLPs (Multi-Layer Perceptrons) \cite{lapedes1988neural}, \cite{zeeshan:Linear}, \cite{zeeshan:single}, \cite{zeeshan:lap}, \cite{zeeshan:cca}, RBF (Radial Basis Function) neural network \cite{casdagli1989nonlinear}, FIR (Finite Impulse Response) neural network \cite{wan1993time}, SVR (Support Vector Regression) \cite{smola2004tutorial}, SOM (Self-Organization Map) \cite{vesanto1997using}, GP (Gaussian Process) echo state machine \cite{chatzis2011echo}, SVESM (Support Vector Echo State Machine) \cite{shi2007support}, RNNs (Recurrent Neural Networks) including NAR (Nonlinear AutoRegressive network) \cite{principe1995dynamic}, Elman networks \cite{zhang1997recurrent}, Jordan networks, RPNN (Recurrent Predictor Neural Networks) \cite{han2004prediction} and ESN (Echo State Network) \cite{rodan2011minimum}. 

\section{Logistic Regression}
\label{section3}
\subsection{Description}
Logistic regression is a widely used algorithm which frequently appears in medical literature \cite{133}. The multivariable method tries to establish a functional relationship between two or more predictor (independent) variables and one categorical outcome (dependent) variable.  The probability of $Y$ occurring is predicted given known values of X (vector containing predictors). The general form of the functional dependence given by the regression could be expressed as per the following formula (1):
\begin{equation}
{\bf X}P({\bf Y}) = (1 + e -(b_{0} + b_{1}x_{1} + ... + b_{n}x_{n}))-1
\end{equation}
where P(Y) is the probability of Y occurring (or, in other words, of Y belonging to a certain class), ${\bf X}_{n}$ are predictor variables and $b_{n}$ are coefficients to be determined by the logistic regression algorithm. The coefficients are estimated by fitting models, based on the available predictors, to the observed data.  

Thus it can also be defined mathematically by relating the probability of some event, E, occurring, conditional on a vector, ${\bf x}$, of explanatory variables, to the vector ${\bf x}$, through the functional form of a logistic edf. Thus
\begin{equation}
\text{p}({\bf x})= \text{Pr}\{\text{E}|{\bf x}|\} = 1/[1 + \text{exp} \{- \alpha - \beta^{'}{\bf x}\}], 
\end{equation} 

where $(\alpha, \beta)$ are ambiguous parameters that are estimated from the data.   

\subsection{Experimental Results \& Discussion}

Logistic regression is always the preferred initial method for describing and testing hypothesis relating to relationships between a categorical/binomial outcome variable and one or more categorical or continuous predictor variables. The binomial outcome variable predicted in this paper is the feature named `ReachedcompositeEndpoint'. The predictive accuracy based on all available 57 attributes, after applying the logistic regression method, is \textbf{73.34} \% which indicates possible overfitting due to the large number of features used. In order to increase the overall predictive accuracy, we used a statistical backward feature selection method \cite{backward} with tenfold cross validation, to identify all significant variables at a confidence interval of 95\%. The list of significant features (at the 95 \% confidence level) was found to be: sex, montreallocationdiagnosis, HBI, Fromlastupdatemontrealmonth, Maxmontrealbehaviour, Plt, AbdominalPainID, LiquidStoolPerDay, MouthUlcer, CountBeforeCEndpoint, MinBeforeCEndpoint, AvgBeforeCEndpoint, AvglogBeforeCEndpoint, MaxBeforeCEndpoint, CountBeforeMontrealIncrease, MaxBeforeMontrealIncrease and AvgBMontrealIncrease. Application of the logistic regression model using these features resulted in an improved predictive accuracy, increasing from \textbf{73.34}\% (with 57 features)  to \textbf{80.02}\% (with 17 features) as shown in Table \ref{logit}. Table \ref{sig} shows the 17 extracted predictive features at 95\% confidence level. 

\begin{table}[h]
\caption{List of identified Predictive features, using a statistical backward feature selection method (at the 95\% Confidence Interval)}

  \begin{tabular}{|p{2.1cm}|p{0.9cm}|p{1.0cm}|p{0.8cm}|p{0.8cm}|p{1.2cm}|}
    \hline
    \textbf{Predictor} & \textbf{Co-eff.}&  \textbf{S.Error} & \textbf{O. Ratio} & \textbf{z-value} & \textbf{p-value}$<=$ 0.0001 \\ \hline
    Sex & 1.15 & .3065 &3.168&3.719& 0.0001\\ \hline
    Montloc. & -0.19 & 0.1134 &0.827&-1.58&0.0469\\ \hline
    HBI & -0.39 & 0.1755 &0.675&-2.31&0.0104\\ \hline
    LastMon & -0.013 & 0.0057 &0.987&-1.308&0.0954\\ \hline  
    MaxMonBeh & 0.58 & 0.1669 &1.791&3.642&0.0001\\ \hline
    Plt & -0.003 & 0.0008 &0.996&-3.52&0.000216\\ \hline
    AbdPaidID & 0.56 & 0.3249 &1.746&5.37&0.0001\\ \hline
    Liqstoolperday & 0.45 & 0.2047 &1.563&2.1817&0.0145\\ \hline
    MouthUlcer & 2.01 & 0.0389 &7.404&51.46&0.0001\\ \hline
    Count & 0.33 & 0.0937 &1.388&3.43&0.0003\\ \hline
    Min & -0.01 & 0.009 &0.996&-4.45&0.0001\\ \hline
    Avg & 0.01 & 0.0016 &1.008&4.98&0.0001\\ \hline
    CompCount & -0.48 & 0.1284 &0.617&-3.89&0.0001\\ \hline
    CompMax & -0.01 & 0.0007 &0.997&-4.29&0.0001\\ \hline
    AvgComp & 1.38 & 0.2978 &4.009&4.662&0.0001\\ \hline
    IncMaxNumCal & -0.01 & 0.0008 &0.998&-2.50&0.00621\\ \hline
    IncAvgCal & -0.94 & 0.3217 &0.387&-2.95&0.0015\\ \hline
 \end{tabular}
\label{sig}

\end{table}

\begin{figure}[ht]
\centering
\includegraphics[width=6cm,height=5cm]{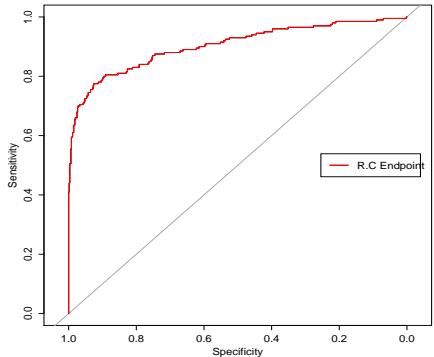}
\caption{Receiving operating characteristic curve for ReachedCompositeEndPoint as predictor of IBD Bowel disease (IBD) using standard Logistic Regression and Backward Feature Selection (CI 95\%)}
\label{fig:swissrol3}
\end{figure}

The confusion matrix shown in Table \ref{accuracy} shows correct classification of 81 patients  (originally labeled `0' and 3 patients labeled `1') and misclassification of 19 patients on an example testing dataset. The classification accuracy on the test dataset is shown in Table \ref{logit}. The receiving operating characteristic of the response variable and the predictor is shown in Figure \ref{fig:swissrol3}

\begin{table}[h]
\caption{Results for Logistic Regression}
\begin{tabular}{|c|c|c|c|}
    \hline
    Algorithm & Training MSE & Testing MSE & Predictive Accuracy \\
    \hline
    Logistic Regression & 0.115 & 0.2002 & 80.02 \% \\
    \hline
\end{tabular}
\label{logit}
\end{table}

\begin{table}
\caption{Confusion Matrix (Backward feature selection using standard Logistic Regression approach) with Test Data}
\centering
\begin{tabular}{|c|c|c|}
\hline
 &{\bf 0}&{\bf 1}\\
\hline
{\bf 0}& {\bf 81}& 19\\
\hline
{\bf 1}&1 & {\bf 3}\\
\hline
\end{tabular}
\label{accuracy}
\end{table}
       
\section{Echo State Network}
\label{section4}
\subsection{Description}
\begin{figure}[h]
\centering
\includegraphics[height=6cm, width=8cm]{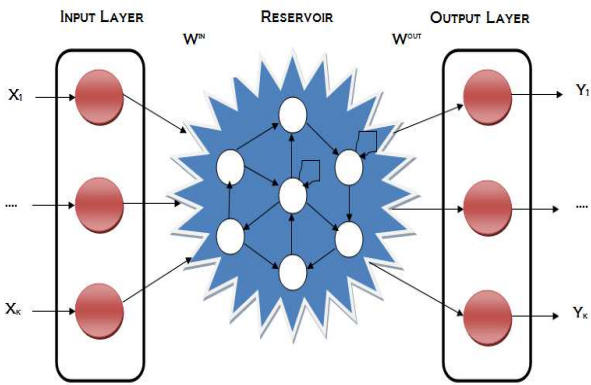}
\caption{\label{echo} Topology of proposed Echo State Network}
\end{figure}

Echo State Networks \cite{reservoir:review} are a popular type of Reservoir Computing are mainly composed of three layers of 'neurons': an input layer, which is connected with random and fixed weights to the next layer, which forms the reservoir. The topology of an echo state network is shown diagrammatically in Figure \ref{echo}. The neurons of the reservoirs are connected with each other through a fixed, random, sparse matrix of weights. Normally only 10\% of the weights in the reservoirs are non-zero. The weights from the reservoir to the output neurons are trained using error descent. Only weights from the reservoirs to the output are trainable. In this paper, we present our initial results which primarily consist of the initial stage of our discovery phase. The prime objective at this initial stage was to find the most appropriate algorithm for maximizing predictive accuracy and further improving the predictive decision making capabilities, to diagnose patients with inflammatory bowel disease. The following state-of-the art ESN \cite{reservoir:review} based recurrent neural network approach has proved the bench mark approach compared to standard logistic regression approach.

The remaining part of this paper describes in detail how the state-of-the-art ESN based recurrent network approach benchmarks against the standard logistic regression approach.

We first define the idea of a reservoir. $W_{in}$ indicates the weight from the $N_u$ inputs ${\bf u}$ to the $N_x$ reservoir units ${\bf x}$, $W$ indicates the $N_x \times N_x$ reservoir weight matrix, and $W_{out}$ indicates the $(N_x +1) \times N_y$ weight matrix connecting the reservoir units to the output units, denoted by ${\bf y}$. Typically $N_x \gg  N_u$. $W_{in}$ is fully connected with the neurons inside the reservoirs and fixed (i.e the weights are non-trainable).  $W$ is also fixed. $W_{out}$ is fully connected and the weights are trainable.

The network dynamics are governed by 
\begin{equation}
{\bf x}(t) = f({\bf W}_{in} {\bf u}(t)+ {\bf W} {\bf x}(t-1) ), \label{reservoirdynamics}
\end{equation}
where $f(.)=\tanh(.)$ and $t$ is the time index. The feed forward stage is given by
\begin{equation}
{\bf y}= {\bf W}_{out} {\bf x}. \label{ff}
\end{equation}

This is followed by supervised learning of the output weights, $W_{out}$. An incremental least mean square method is used for online learning as follows:
\begin{equation}
{\bf W}_{out}= {\bf W}_{out} + \eta ({\bf y}_{target} - {\bf y}){\bf x}^T,
\label{outputtrained}
\end{equation}
where $\eta$ is a learning rate (step size) and ${\bf y}_{target} $ is the target output corresponding to the current input.

\subsection{Experimental Results \& Discussion}
A ten-fold cross validation approach was adopted to avoid overfitting. The network was optimised for the number of hidden neurons, and choice of learning parameters. The weights of the reservoir were incrementally aligned by re-setting their spectral radius, following a random initialization of the reservoir weights. A regularization co-efficient was also introduced in the cost function to avoid over-fitting, by penalizing high coefficients for predictors, which also helped stabilize the estimates. The size of the reservoir was empirically set to 150. The nominal selection of neurons inside the reservoir was found to be important for maximizing predictive accuracy.

As shown in Table \ref{crossvalall}, the proposed echo state network significantly outperformed logistic regression, creating a new benchmark accuracy of 96.04 \%. One interesting point to note from Table \ref{crossvalall} is the decrease in predictive accuracy when all 57 independent variables were used for learning, which is also the case for the Logistic regression model. Specifically, the ESN predictive accuracy was found to increase from 86.78\% to 96.04 \% by using the 17 most significant features at a confidence interval of 95 \%; whereas there was a slight decrease in the predictive accuracy when selecting the 15 most significant features at a confidence interval of 99\%.  

The average error as shown in Figure \ref{fig:swissrol3} was initially noted a bit higher by using all independent features even with ESN shown in Table \ref{crossvalall}; which was then further decreased by performing ten-fold cross validation with standard backward selection criteria; and extracting most significant features from the input dataset both at 95 \% and 99 \% confidence interval. This procedure of feature selection proved extremely useful with the standard recurrent neural network ESN based approach compared to the standard logistic regression approach explained previously in section \ref{section3}. The list of extracted highly significant features was the exact same as those in section \ref{section3}. The receiver operative characteristic curve which clearly shows the relationship between sensitivity and specificity along with average square error; drastically decreases to 0.0396 and 0.0400 at 95 \% and 99 \% confidence intervals is shown in Figure \ref{fig:swissrol6} and Figure \ref{fig:swissrol8} respectively. 

\begin{table}
\caption{ESN \cite{reservoir:review} Results}
\begin{tabular}{|p{1cm}|p{1cm}|p{0.8cm}|p{0.8cm}|p{1.4cm}|p{0.6cm}|p{0.5cm}|}
\hline
 Independent variables  & Dependent variable &Avg MSE &Standard Deviation&Variance& Mean Accuracy &CI \%\\
\hline
57&1&0.3179&0.3637&0.1322& 86.78\% &-\\
\hline
17&1&0.0396&0.0157&$2.4799e^{-004}$& 96.04\%&95\%\\
\hline
15&1&0.0400&0.0099&$9.8815e^{-004}$ &96\%&99\%\\
\hline
\end{tabular}
\label{crossvalall}
\end{table}

\section{Conclusion}
\label{section5}
In this paper we have introduced a novel ESN based predictive model, for retrospective FC analysis, and benchmarked it against the conventional Logistic regression model. Our model outperformed the logistic regression model and was found to be more accurate.

Future work will include extending the initial predictive analysis in this paper, by applying our novel ESN approach to the following cases, and bench marking against conventional logistic regression analysis - to identifying factors predictive of reaching: (1) secondary end points (2) advancement in montreal behaviour (3) surgical resection (4) hospitalisation for flares

Finally, we will apply and comparatively evaluate both single layered \cite{reservoir:review}, and our newly proposed Multiple layer Echo state network model \cite{zkmalik} on this massively unstructured retrospective data set. The latter model has already proven to be more accurate in reducing error and in computational competitiveness.\cite{zkmalik}.



%
\IEEEpeerreviewmaketitle

\onecolumn
{
\begin{figure}[ht]
\centering
\begin{minipage}[b]{0.38\linewidth}
\includegraphics[width=6cm,height=5cm]{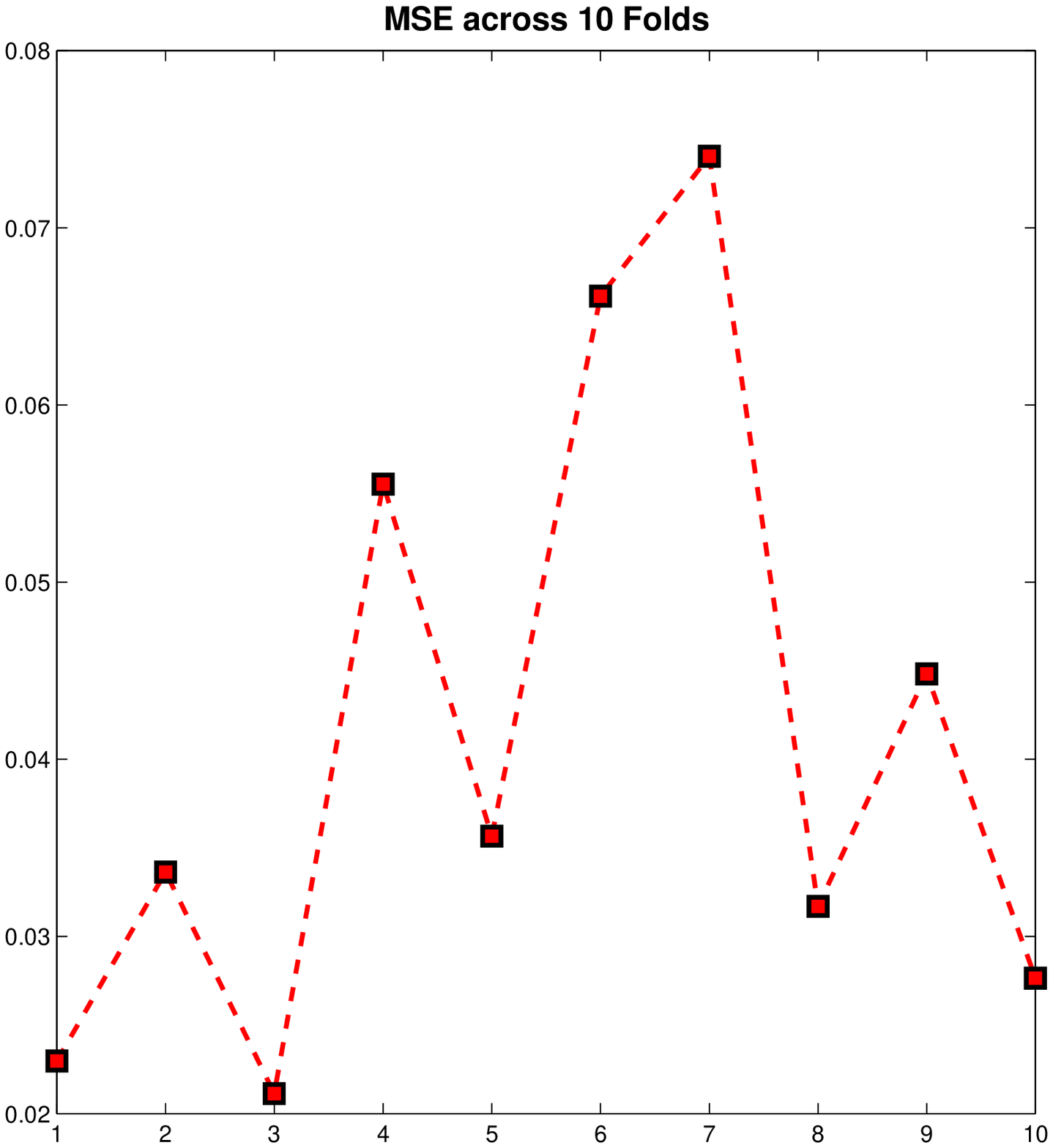}
\caption{Average Mean Square Error at 95\% CI}
\label{fig:swissrol6}
\end{minipage}
\begin{minipage}[b]{0.38\linewidth}
\includegraphics[width=6cm,height=5cm]{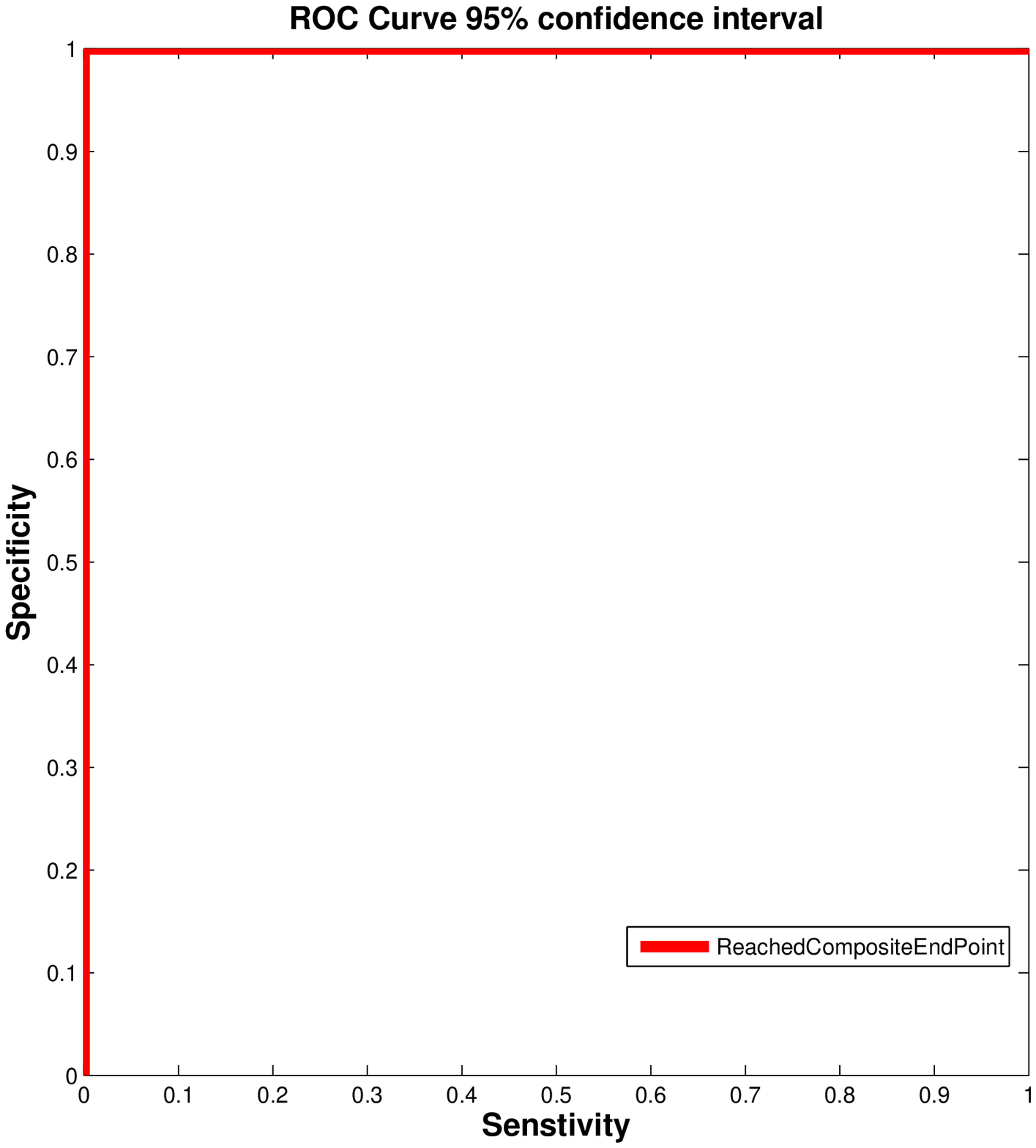}
\caption{Receiver Operating Characteristic (ROC) curve}
\label{fig:swissrol7}
\end{minipage}
\end{figure}
\begin{figure}[ht]
\centering
\begin{minipage}[b]{0.38\linewidth}
\includegraphics[width=6cm,height=5cm]{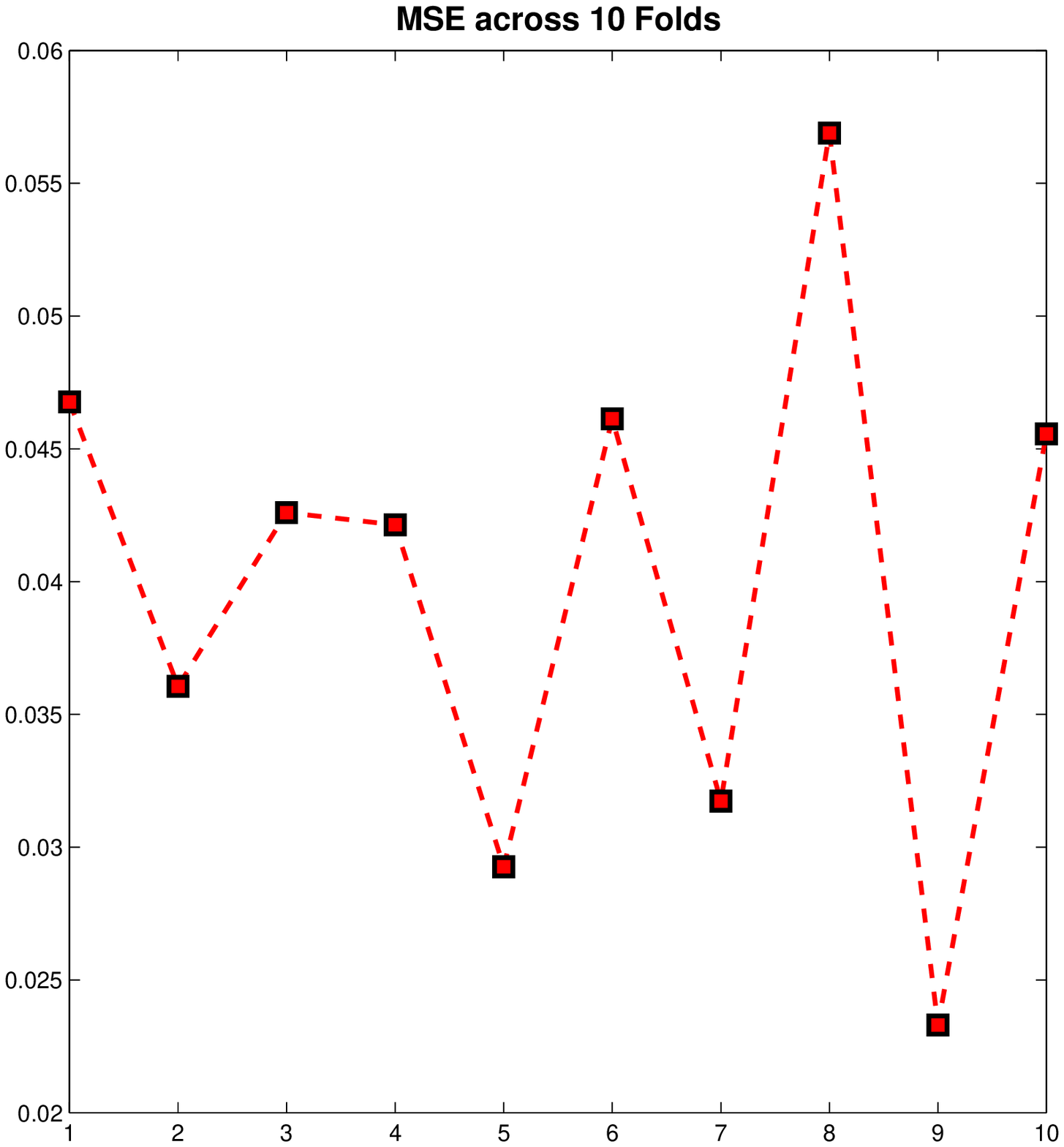}
\caption{Average Mean Square Error at 99\% CI}
\label{fig:swissrol8}
\end{minipage}
\begin{minipage}[b]{0.38\linewidth}
\includegraphics[width=6cm,height=5cm]{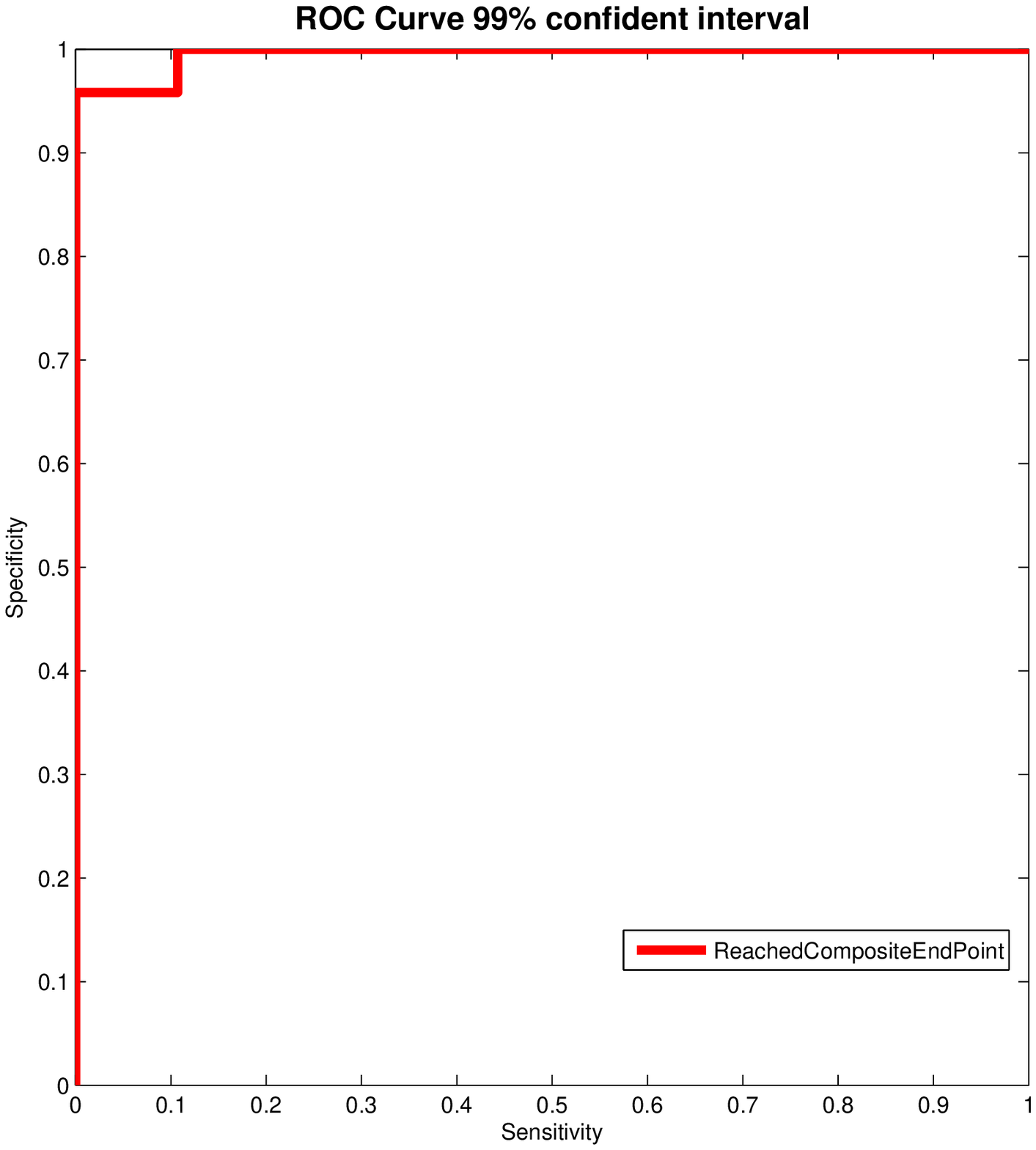}
\caption{Reciever Operating Characteristic (ROC) curve}
\label{fig:swissrol9}
\end{minipage}
\end{figure}
}

\end{document}